\documentclass[letterpaper, 10pt, conference]{ieeeconf}      

\IEEEoverridecommandlockouts                              

\title{\LARGE \bf
I2P-Rec: Recognizing Images on Large-scale Point Cloud Maps through Bird's Eye View Projections
}

\author{Shuhang Zheng$^{\dagger}$, Yixuan Li$^{\dagger}$, Zhu Yu, Beinan Yu, Si-Yuan Cao, Minhang Wang, Jintao Xu,\\ Rui Ai, Weihao Gu, Lun Luo$^*$, Hui-Liang Shen
\thanks{$^{\dagger}$ contributed equally.}
\thanks{$*$ corresponding author.}
\thanks{Shuhang Zheng is with the College of Optical Engineering, Zhejiang University, Hangzhou 310027, China.  {\tt zhengsh@zju.edu.cn}.}%
\thanks{Yixuan Li, Zhu Yu, Beinan Yu are with the College of Information Science and Electronic Engineering, Zhejiang University. {\tt{\{yixuanli, yu\_zhu\}@zju.edu.cn, mr\_vernon@hotmail.com}.}}
\thanks{Si-Yuan Cao is with the Ningbo Research Institute, Zhejiang University, Ningbo 315100, China. \tt{karlcao@hotmail.com}.}
\thanks{Minhang Wang, Jintao Xu, Rui Ai, Weihao Gu are with Homo.AI Technology Co., Ltd, Beijing 100192, China. \tt{\{wangminhang, xujintao, airui, guweihao\}@haom.ai}}
\thanks{Lun Luo is with  the College of Information Science and Electronic Engineering, Zhejiang University, and also with Homo.AI Technology Co., Ltd. \tt{luolun@zju.edu.cn}.}
\thanks{Hui-Liang Shen is with the College of Information Science and Electronic Engineering, Zhejiang University, and also with the Ningbo Research Institute, Zhejiang University, Ningbo 315100, China. \tt{shenhl@zju.edu.cn}.}
}

\usepackage{graphicx}
\usepackage{multirow}
\begin{document}

\maketitle
\thispagestyle{empty}
\pagestyle{empty}

\begin{abstract}
Place recognition is an important technique for autonomous cars to achieve full autonomy since it can provide an initial guess to online localization algorithms. Although current methods based on images or point clouds have achieved satisfactory performance, localizing the images on a large-scale point cloud map remains a fairly unexplored problem. This cross-modal matching task is challenging  due to the difficulty in extracting consistent descriptors from images and point clouds. In this paper, we propose the I2P-Rec method to solve the problem by transforming the cross-modal data into the same modality. Specifically, we leverage on the recent success of depth estimation networks to recover point clouds from images. We then project the point clouds into Bird's Eye View (BEV) images. Using the BEV image as an intermediate representation, we extract global features with a Convolutional Neural Network followed by a NetVLAD layer to perform matching. The experimental results evaluated on the KITTI dataset show that, with only a small set of training data, I2P-Rec achieves recall rates at Top-1\% over 80\% and 90\%, when localizing monocular and stereo images on point cloud maps, respectively. We further evaluate I2P-Rec on a 1 km trajectory dataset collected by an autonomous logistics car and show that I2P-Rec can generalize well to previously unseen environments.

\end{abstract}

\section{INTRODUCTION}
Localization is an essential module for autonomous cars to perform perception, planning, and navigation \cite{cadena2016past}. A commonly used localization approach is to match the Light Detection And Ranging (LiDAR) scans with a pre-built point cloud map \cite{lu2019l3}. This method can exploit the high precision and rich geometric information of point cloud maps. However, the method is hard to deploy commercially due to the high price and large volume of LiDAR scanners, although the maps have been widely used in fields such as autonomous driving. As an alternative, camera is more competitive because it is much cheaper, smaller, and can provide rich semantic information. Thus, the question comes out naturally that whether we can localize the images in point cloud maps. In recent years, some methods \cite{feng20192d3d, chen2022i2d} try to tackle this problem by matching the images with point clouds based on the local features. A main limitation of these methods is that they need to know the rough position of a camera to give an initial guess, which itself is a problem to be solved. To fill this gap, we work on estimating the rough locations of images on a large-scale point cloud map without other prior information.
 
\begin{figure}[t]
      \centering

      \includegraphics[width=3in]{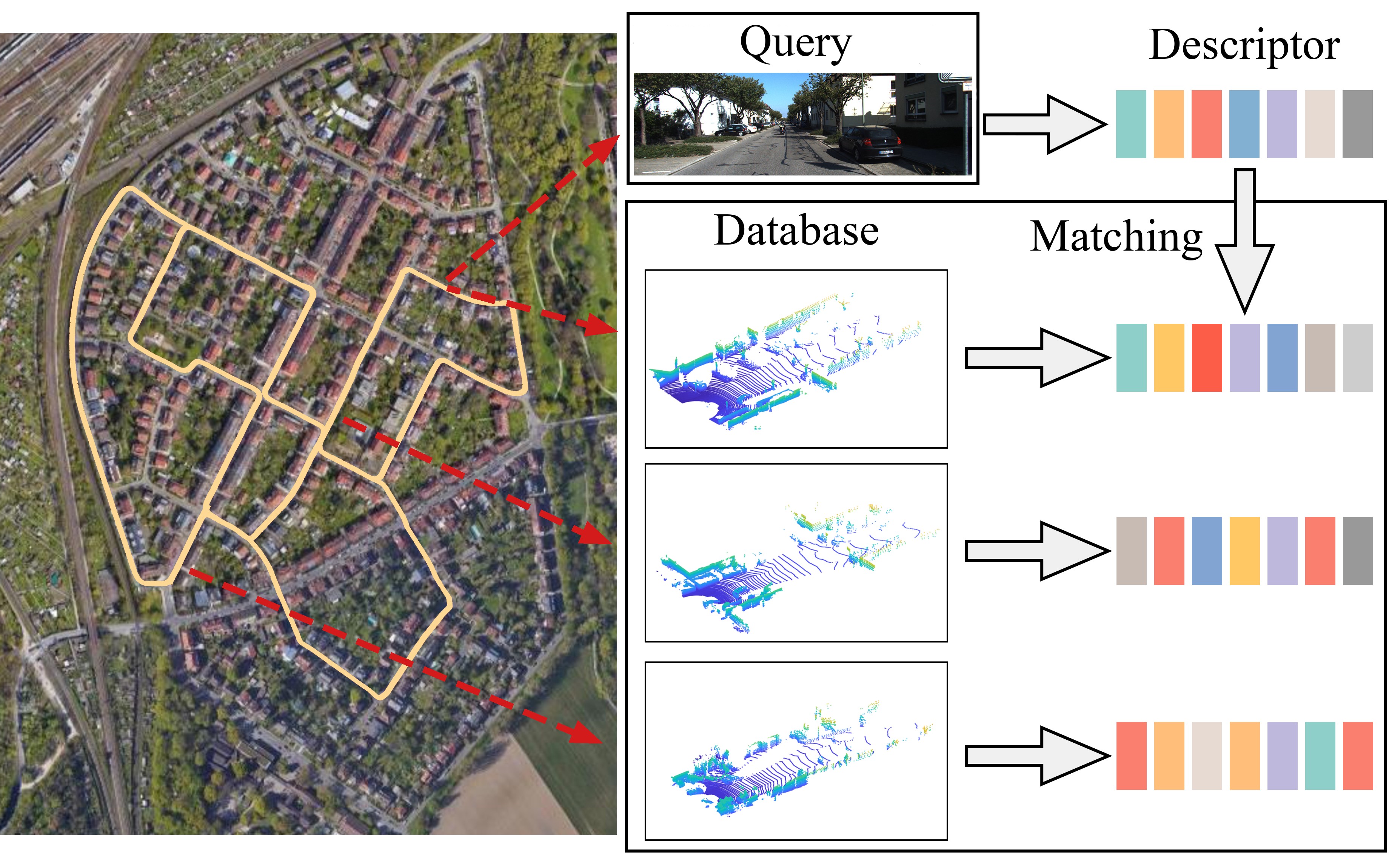}

      \caption{Recognizing places of images on large-scale point cloud maps. Global descriptors are extracted from both images and point clouds. Place recognition is performed by searching the database for point clouds with the minimum descriptor distance to the query image.}
      \vspace{-5mm}
      \label{fig: cross_modal_vpr}
   \end{figure}
   
The rough location estimation problem can be viewed as a place recognition problem, which is to search through a pre-built database to retrieve the closest frame to a query, as shown in Fig. \ref{fig: cross_modal_vpr}. Lots of place recognition methods \cite{angelina2018pointnetvlad,liu2019lpd,bvmatch2021,du2020dh3d,luo2022lidar,2012Bags,arandjelovic2016netvlad} have been proposed in cases where queries and databases share the same data modality, such as images or point clouds. However, how to localize images in a point cloud database remains a challenging and unsolved problem due to the following reasons. First, it is hard to model the similarity between the data from two different types of sensors. The LiDAR obtains depth information by calculating the laser time of flight (TOF), while the camera gathers photometric information through the response of CMOS to ambient light. There is little explicit correlation between the data of these two modalities. Second, it is difficult to design a unified feature extractor as the format of data varies. LiDAR scans are often organized as unordered points, whereas the images are in the form of regular grids. 

Cattaneo et al. \cite{cattaneo2020global} introduces a method to perform visual place recognition on LiDAR maps. They exploit a 2D and a 3D Deep Neural Network (DNN) to extract features from images and point clouds respectively, and create a shared embedding space between images and the LiDAR map. However, without special designs to deal with the modality difference, this method may have weak generalization ability. In this work, we propose our approach called I2P-Rec to solve the problem by modality transformation. We first leverage the monocular and stereo depth estimation methods \cite{godard2019digging,chang2018pyramid,xie2023darkmim} to recover point clouds from the input images. This procedure unifies the modality of data used for place recognition. Inspired by recent works on LiDAR-based place recognition methods \cite{bvmatch2021,luo2023}, we project the point clouds from both the query images and database into Bird's Eye View (BEV) images. We then use a modern Convolutional Neural Network (CNN) for global feature extraction. We evaluate our method on both the KITTI dataset \cite{kitti} and a dataset collected by ourselves, demonstrating that, 

1) with only a small amount of training data, I2P-Rec can effectively localize images on large-scale point cloud maps;

2) I2P-Rec can generalize well to unseen environments; 

3) I2P-Rec benefits from the accuracy of depth estimation and can achieve higher recall rates with better depth estimation algorithms.


We believe our method can serve as a good baseline for the task of visual place recognition on point cloud maps. With a simple yet effective design, we hope our method can draw the attention of the robotics community to the problem of cross-modal place recognition, and lead to the emergence of more excellent algorithms.

\section{RELATED WORKS}
In this section, we review the place recognition methods that work on single-modal data. We then discuss the recent development of the approaches to visual place recognition on point cloud maps. 
\subsection{Place Recognition Based on Single-modal Data} 
Camera is a widely used sensor for place recognition. Benefiting from the local image features \cite{2004Distinctive,rublee2011orb} and the bag-of-words techniques \cite{lazebnik2006beyond}, a number of image-based place recognition methods \cite{2012Bags,2017ORB,pronobis2006discriminative,lowry2015visual, angeli2008fast} have been developed. These methods are proposed based on an assumption that similar structures in the environment lead to similar local feature distributions. They first learn the feature distribution by clustering  the local features into clusters, and then compute a feature vector according to the distance of each local feature to cluster centers. Most subsequential place recognition methods try to improve performance by designing better local feature extractors and feature aggregators. For example, NetVLAD \cite{arandjelovic2016netvlad} extracts local features by CNN and designs a network to learn the cluster centers. Patch-NetVLAD \cite{hausler2021patch} follows NetVLAD and leverages the strengths of both the global and local features to generate more distinct patch-level features. 

In recent years, LiDAR-based place recognition has become a hot topic due to the ability of LiDARs to acquire accurate depth information and its robustness to illumination changes and view variations.  Following the setup of NetVLAD \cite{arandjelovic2016netvlad}, PointNetVLAD \cite{angelina2018pointnetvlad} uses the point cloud encoder, i.e. PointNet \cite{qi2017pointnet} for local feature extraction, and then generates global features with NetVLAD. To take full advantage of contextual information, PCAN \cite{zhang2019pcan} introduces the point contextual attention network to extract task-relevant features. However, both PointNetVLAD and PCAN cannot capture local geometric structures due to their independent treatment for each point. Thus, the following methods mainly focus on exploiting neighborhood information \cite{liu2019lpd,soe,mickloc3dv2}. On the other hand, some methods project point clouds into images and perform recognition with features from the images. Kim et al. \cite{kim2018scan}  propose the scan context descriptor by projecting point clouds into an ego-centric coordinate system built by partitioning the ground space into bins according to both azimuthal and radial directions. OverlapNet \cite{chen2020overlapnet, OverlapTransformer} adopts a siamese network to learn the overlap between a pair of range images. BVMatch \cite{bvmatch2021} projects point clouds into BEV images and introduces the BVFT feature for global feature extraction and pose estimation. It shows that the BEV is quite effective representation and BEV projection-based methods \cite{bvmatch2021,luo2022lidar} achieve state-of-the-art performance in terms of retrieval recall, robustness, and generalization ability. In this work, we utilize BEV projection to build a bridge between images and LiDAR maps. 

\subsection{Visual place recognition on point cloud maps}
Although place recognition based on single-modal data has been studied a lot, how to recognize images on a point cloud map remains a challenging problem. One straightforward solution is jointly training a 2D CNN for images and a 3D DNN for point clouds to create shared embeddings \cite{cattaneo2020global}. However, this approach does not generalize well to unseen environments. \cite{bernreiter2021spherical} proposes to project images and point clouds into unit spheres and extract features through sphere CNN \cite{esteves2018learning, cohen2018spherical}. This method requires multiple images as input but our method only uses a monocular image or a pair of stereo images. Some methods \cite{feng20192d3d, chen2022i2d} detect local features from images and point clouds and match them for place retrieval. These methods can only be deployed in a local area and cannot be extended to a large-scale environment well.

\begin{figure*}[!thpb]
      \centering
      \includegraphics[width=6in]{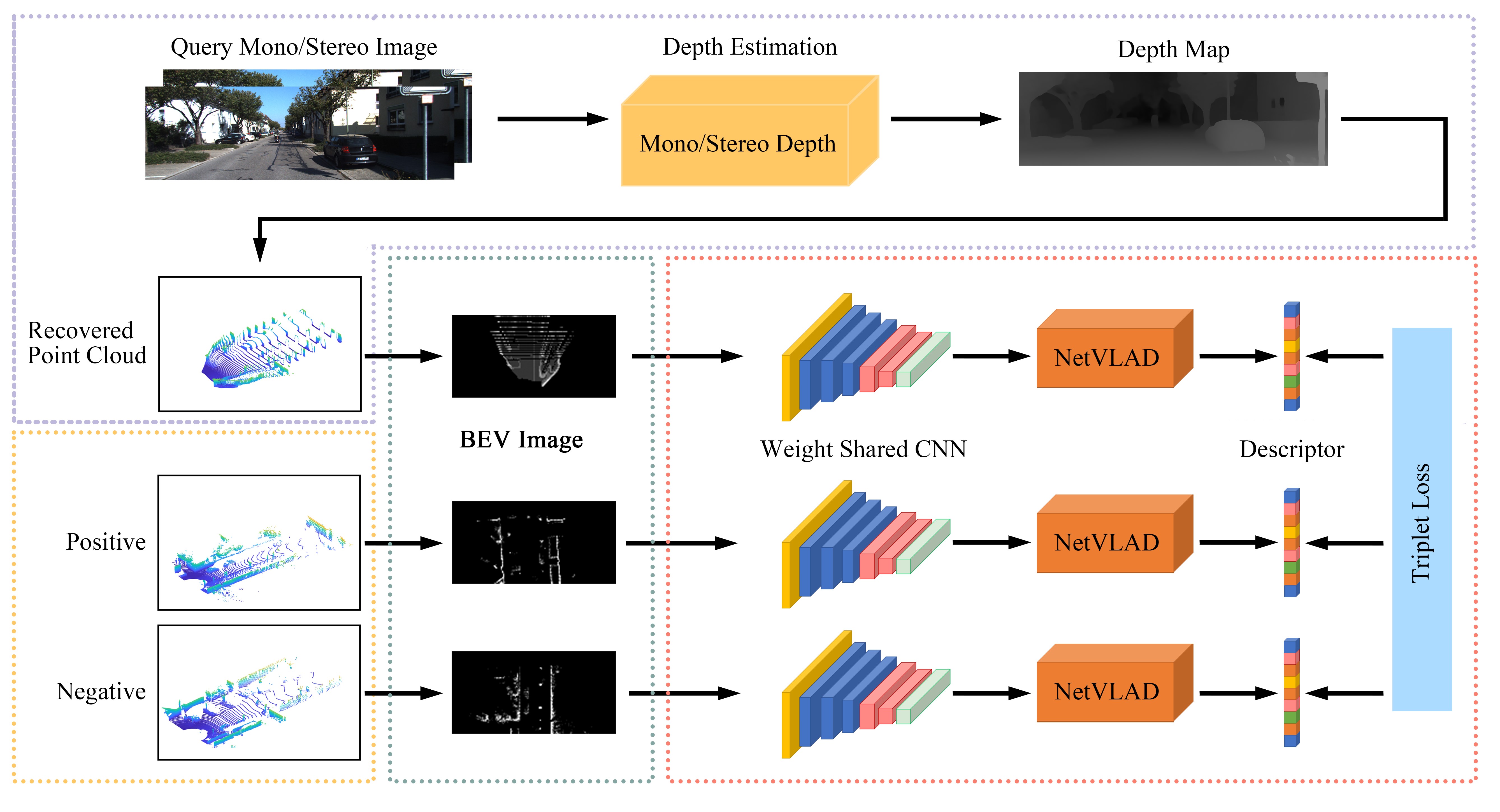}
      \caption{The training framework of the proposed I2P-Rec. Query images are converted to point clouds via a depth estimation module. Then, both point clouds from queries and the database are projected to BEV images. Based on the BEV images, global features are extracted by a CNN and a NetVLAD layer. Triplet loss is adopted for supervision. }
      \label{fig: trainging}
\end{figure*}
\section{METHOD}
The visual place recognition on point cloud maps can be defined as, given an image, looking into a point cloud database with known poses and retrieving the one representing the same place of the query image. We propose I2P-Rec to solve the problem. In this method, we first extract global descriptors from a set of point clouds collected in a designated area. We build the database with the point clouds and their corresponding descriptors. Then, we generate global descriptors of the query images and retrieve the very frames with the minimum feature distance to the queries from the database. 

The key of our I2P-Rec is to generate global descriptors that can model the appearance similarity between images and point clouds. In this process, we have to take special consideration of the modality difference between the queries and the database. To this end, we transform both the images and point clouds into BEV images for feature extraction and perform matching on the basis of the BEV representation. In the following sections, we first introduce point cloud recovery from the query images via depth estimation methods. We then explain the process of BEV-image generation. Finally, we detail our network architecture. 
\subsection{Depth Estimation}
According to the camera setup, depth estimation methods can be divided into two types, monocular methods using a single camera, and stereo methods using a pair of cameras. Monocular methods usually recover the depth directly from a single image and predict scale information of monocular depth with the training based on depth supervision or consistency between stereo images \cite{monosurvey}. On the other hand, stereo methods first predict a disparity map $D(u,v)$, and then generate the depth map from it. The depth of a pixel at position $(u,v)$ can be derived as
\begin{equation}
        \mathrm{depth}(u,v) = \frac{f_U*b}{D(u,v)},
        \label{eq1}
\end{equation}
where $b$ is the horizontal offset between a pair of cameras (i.e., baseline) and $f_U$ is the horizontal focal length. 

After obtaining the depth map, we back-project each pixel to a 3D point. We establish the coordinate system according to the right-hand rule, with the x-axis pointing to the right, the y-axis pointing forward, and the z-axis pointing upward. In this coordinate system, the x-y plane is the ground plane. For each pixel at position$(u,v)$ with a depth value of $depth(u,v)$, we obtain the corresponding 3D point $(x_i, y_i, z_i)$ by 
\begin{equation}
        [x_i, y_i, z_i]^\mathrm{T} = \textbf{R}^{-1}\left(\textbf{K}^{-1}\left[u, v, \mathrm{depth}(u,v)\right]^\mathrm{T}-\textbf{t}\right),
        \label{eq2}
\end{equation}
where \textbf{K}  is the camera intrinsic matrix, while \textbf{R} and \textbf{t} correspond to the extrinsic rotation matrix and translation vector respectively. $i$ stands for the point counts.  

Our I2P-Rec is agnostic to different depth estimation algorithms. In general, the more accurate the depth estimation is, the better performance our method can achieve. In the experiment section, we will evaluate our method with both monocular and stereo depth estimation algorithms.

\subsection{Projecting Point Clouds to BEV Images. }
Using BEV images as an intermediate representation has been adopted by some LiDAR-based place recognition methods \cite{bvmatch2021, luo2023}. In this work, we generate BEV images following BVMatch \cite{bvmatch2021}. We first grid the ground plane into squares, each with a side length of 0.4m, and then project the point clouds perpendicularly to the ground. We count the points in each grid and take the normalized number of points as values for each pixel in a BEV image. 

Another problem arises that the size of BEV image generated from images and LiDAR scans may be different though representing the same areas. This is because the cameras and LiDAR scanners have different view ranges. Usually, multi-beam LiDAR sensors own a 360-degree field of view, whereas the camera only senses the cone-shaped area in the front. To deal with the situation, we crop the point clouds from both images and LiDAR scanners into a window with the x-axis limited to $[$-25 meters, 25 meters$]$, y-axis to $[$0 meters, 50 meters$]$, and z-axis to $[$-5 meters, 5 meters$]$. As we perform projection on the x-y plane, the resulting BEV image is of size $125\times 125$.

\subsection{Network Architecture}
We use a network combined with a CNN network and a NetVLAD layer \cite{arandjelovic2016netvlad} in our I2P-Rec to generate global descriptors from the BEV images. The details of the network architecture and the training process are shown in Fig. \ref{fig: trainging}. During the training process, we find the positive and negative matches for the query image according to ground truth distance. Then we transform these data into BEV images. We extract local features of the images with weight-shared CNN and generate global descriptors with the NetVLAD. For training supervision, we adopt the lazy-triplet loss which can be written as
    \begin{equation}
        \mathcal{L}_{triplet}=\max_j([m+\delta_{pos}-\delta_{negj}]_+)
        \label{eq3}
    \end{equation}
where $[...]_+$ denotes the hinge loss, $m$ is the constant margin, $\delta_{pos}$ is the feature distance between the query and its positive frame, and $\delta_{neg}$ is the feature distance between the query and its negative frames.

In the implementation, we use the ResNet-34 \cite{he2016deep} as our CNN backbone. we set the triplet margin m=0.5, and the number of clusters of NetVALD as 64. Moreover, in the training stage, we apply 2 positives and 10 negatives for loss calculation. We also adopt the hard mining strategy \cite{shrivastava2016training} following NetVLAD after the first 10 training epochs. 

\section{EXPERIMENTS}
In this section, we evaluate the performance of our I2P-Rec for visual place recognition on point cloud maps. We compare it with the baseline method \cite{cattaneo2020global}, and test it with both monocular depth estimation methods and stereo algorithms. 

   \begin{table}[t]
\caption{Partition of the dataset.}
\vspace{-3mm}
\label{table1}
\begin{center}
\setlength{\tabcolsep}{1.5mm}{
\begin{tabular}{c c c c c c c c}
\hline
&Train &{Val} &\multicolumn{4}{c}{Test}  \\
\hline
                     sequence & 00 & 00 & 02 & 05 & 06 & 08  \\
frame&0-3000 & 3200-4540 & 0-4660 & 0-2760 & 0-1100 & 0-4070\\
\hline
\end{tabular}
\vspace{-3mm}
}
\end{center}
\end{table}

\begin{table}[b]
\vspace{-3mm}
\caption{The recall@1 of monocular visual place recognition methods on the KITTI dataset.}
\vspace{-3mm}
\label{table2}
\begin{center}
\begin{tabular}{c c c c c c}
\hline
\multicolumn{1}{c}{Sequence} & \multicolumn{1}{c}{00} & \multicolumn{1}{c}{02}&\multicolumn{1}{c}{05}&\multicolumn{1}{c}{06}&\multicolumn{1}{c}{08}\\
\hline
Baseline Method \cite{cattaneo2020global} & 58.0 & 4.0 &  10.6 &  22.3 & 5.3 \\
MIM-Points & 53.3 & 6.9 & 14.8  &  20.0 & 12.5 \\
MIM-I2P & \textbf{74.3} & \textbf{46.3} & \textbf{49.9}  & \textbf{41.9}  & \textbf{42.1} \\
\hline
\end{tabular}
\end{center}
\end{table}

\subsection{Setup}
\subsubsection{Dataset}
We conduct experiments on the KITTI dataset \cite{kitti}. KITTI contains a large number of synchronized and calibrated point clouds and images and is widely used to verify autonomous driving-related algorithms. We use the 00, 02, 05, 06, and 08 sequences of its Odometry subset for evaluation. As shown in Table \ref{table1}, we take the images and point clouds of 0-3000 frames in sequence 00 for training, 3200-4540 frames in sequence 00 for validation, and the rest of the sequences for testing. 

To further validate the generalization ability of I2P-Rec, we construct the MAPO dataset. This dataset was built by a rotating LiDAR of 32 beams and a single camera mounted a mobile robot. It covers a trajectory length of over 1 km in MAPO, Beijing City of China. There are 3770 frames of image in the dataset and we use the frames 0 to 1000 for training and the frames 1001 to 3770 for testing.


\begin{table*}[t]
\caption{The recall@1 and recall@1\% of all the methods on the KITTI dataset.}
\vspace{-2mm}
\label{table3}
\begin{center}
\setlength{\tabcolsep}{4mm}{
\begin{tabular}{c c c c c c c c c c c}
\hline
\multicolumn{1}{c}{Sequence} & \multicolumn{2}{c}{00} & \multicolumn{2}{c}{02}&\multicolumn{2}{c}{05}&\multicolumn{2}{c}{06}&\multicolumn{2}{c}{08}\\
recall & @1 & @1\% & @1 & @1\% & @1 & @1\% & @1 & @1\% & @1 & @1\% \\
\hline
MIM-Points & 53.3 & 71.6 & 6.9 & 57.1 & 14.8 & 68.3 & 20.0 & 44.4 & 12.5 & 68.6 \\
PSM-Points & 58.0&72.3 & 8.2&53.6 &  16.0&71.1 &  32.1&50.0 & 14.7&75.0 \\
LEA-Points & 58.8&72.6 & 9.1 & 61.3&21.8&80.0  & 31.6 &71.6 & 18.0&76.3 \\
\hline
MIM-I2P & 74.3 & 88.6 & 46.3 & 89.7 &  49.9 & 89.1 &  41.9 & 81.5 & 42.1 & 88.4 \\
PSM-I2P & 81.7 & 97.4&47.6&92.9 & 63.8&94.1  &  52.1&88.5 & 53.5&92.1 \\
LEA-I2P & \textbf{92.6} & \textbf{99.7} & \textbf{77.0} & \textbf{98.4}&\textbf{83.4}  & \textbf{98.8}&\textbf{55.5}  & \textbf{91.5}&\textbf{69.4}&\textbf{96.4} \\
\hline
\end{tabular}
}
\vspace{-3mm}
\end{center}
\end{table*}
\begin{figure*}[t]
      \centering

      \includegraphics[width=7in]{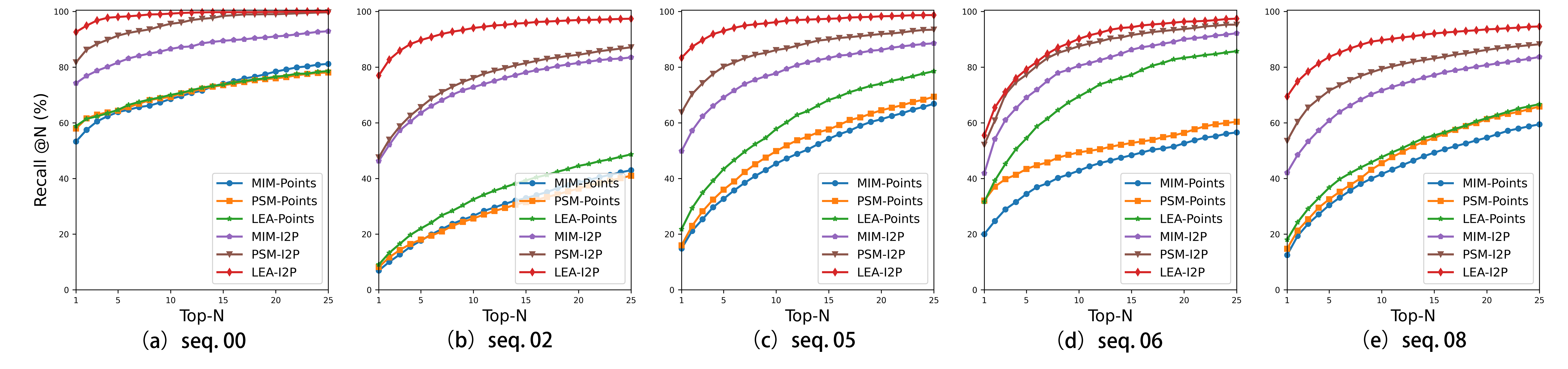}
      \vspace{-5mm}
      \caption{The Top-N recall rate on the KITTI dataset.}
      
      \label{fig_recalltopn}
\end{figure*}
%
%
\subsubsection{Method Setup}
We evaluate our method with different monocular and stereo depth estimation algorithms. For monocular algorithms, we use MIM-Depth \cite{xie2023darkmim} which achieves state-of-the-art depth estimation performance. For stereo algorithms, we adopt PSM-Net \cite{chang2018pyramid} and LEA-Stereo \cite{cheng2020stereo}. For all these methods, we use their open sourced implementations and pre-trained models released on the websites\footnote{https://github.com/JiaRenChang/PSMNet, https://github.com/SwinTransformer/MIM-Depth-Estimation, https://github.com/XuelianCheng/LEAStereo}. 

For clarity, we denote our method with different depth estimation algorithms as \textit{name of depth estimation algorithm - I2P}, i.e., MIM-I2P, PSM-I2P, and LEA-I2P in the following experiments. 

For a more comprehensive evaluation, we also build a place recognition solution that uses point clouds as an intermediate representation. Specifically, we recover point clouds from images via depth estimation and we extract global descriptors from the clouds with PointNetVLAD \cite{angelina2018pointnetvlad}. We denote this method using different depth estimation algorithms as MIM-Points, PSM-Points, and LEA-Points. 

\subsubsection{Evaluation Metrics} 
We adopt the recall at Top-1 and the recall at Top-1\% as the evaluation metrics. When performing the matching, we extract global descriptors from query images and retrieve Top-N nearest matches. We regard a match between a query image and the point cloud as true positive when their geometric distance is less than a threshold of $t=10$ meters. Recall at Top-N is computed as the proportion of correct predictions among all positive instances in a dataset considering the Top-N prediction for each sample. 
\subsection{Monocular Visual Place Recognition}

We evaluate the place recognition performance based on monocular images. We compare our method with the baseline method \cite{cattaneo2020global} that adopts 2D and 3D DNNs to  extract features from query images and point clouds, respectively. Table \ref{table2} shows the recall@1 of the methods on the KITTI dataset. It can be seen that the baseline method achieves modest recall on the validation sequence but can hardly generalize to the test sequences. It is unexpected that, even with the modality transformation, MIM-Points show no obvious superiority to the baseline method. In contrast, MIM-I2P achieves the best recall on both validation and test sequences and surpasses other methods with large margins. This result demonstrates the importance of BEV representation in our method.

\subsection{Better Depth Estimation, Finer Place Recognition}
We further test our method using more depth estimation algorithms, among which the estimation accuracy of MIM-Depth, PSM-Net, and LEA-Stereo increases sequentially. Table \ref{table3} shows the recall rates at Top-1 and Top-1\% on KITTI. It can be seen that our I2P method with the BEV representation outperforms the points-based methods, and it shows better generalization ability. 
    \begin{figure}[!]
      \centering
      \includegraphics[width=2in]{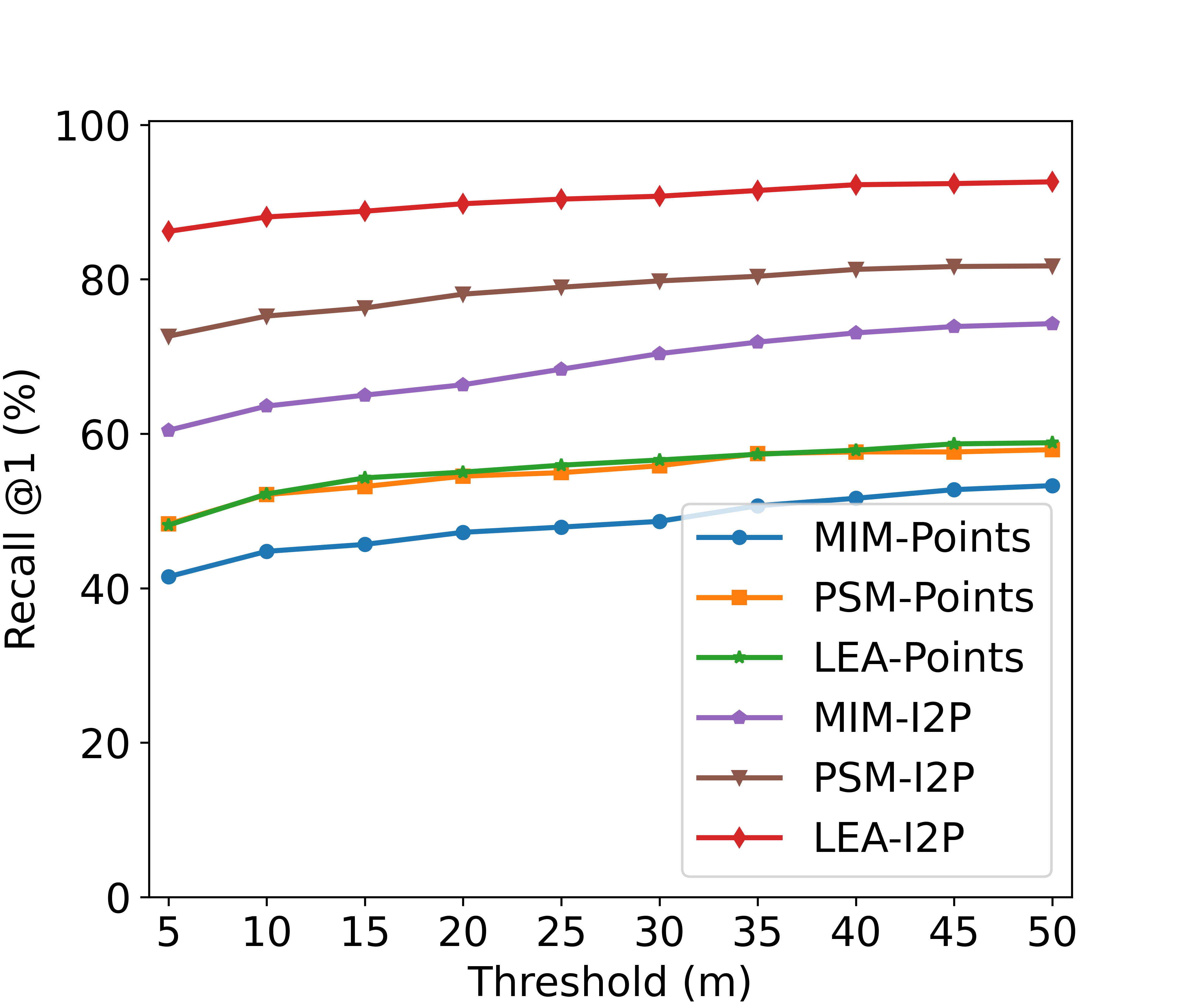}
      \vspace{-2mm}
      \caption{The Top-1 recall with respect to different thresholds on seq 00.}
      \vspace{-5mm}
      \label{fig_threshold}
   \end{figure}
It is noted that, for all the I2P methods and points-based methods, the recall rates are positively correlated with the depth estimation accuracy. However, the performance of the I2P methods improves significantly as the accuracy of depth estimation increases, while the enhancement in points-based methods is not obvious. For a more comprehensive demonstration of the performance of our I2P method, we show the recall rate at Top-N in Fig. \ref{fig_recalltopn}. It can be seen that I2P-Rec achieves higher recall rates on all the sequences.

We also explore the sensitivity of the performance of different methods to the ground truth thresholds. Fig. \ref{fig_threshold} shows the recall@1 when the threshold varies from 5m to 50m. It can be seen that our I2P method can maintain a high recall@1 regardless of the varying thresholds.

\subsection{Performance on MAPO dataset}
\begin{figure}[!]
	\centering
	\includegraphics[width=2in]{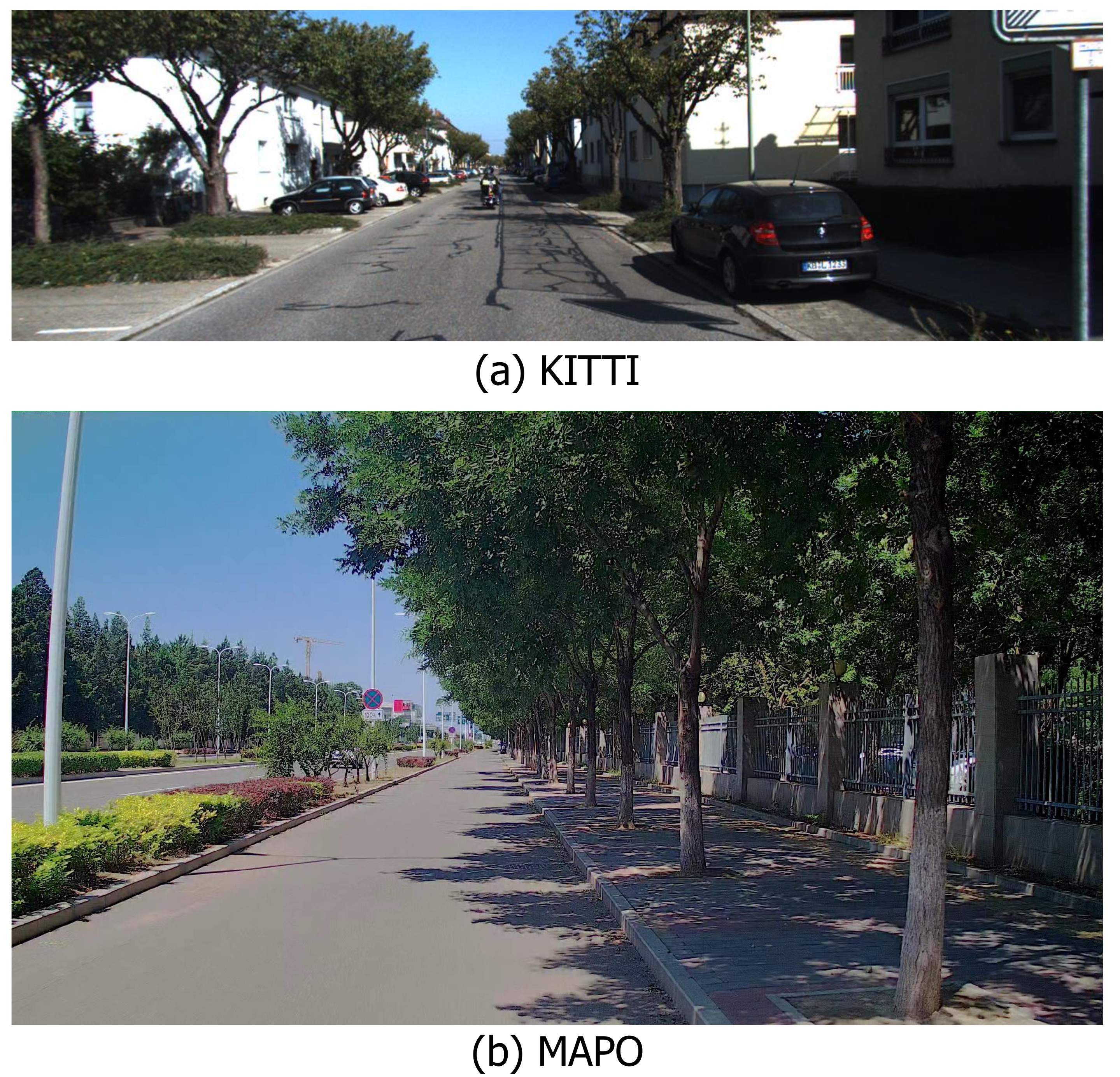}
	\vspace{-3mm}
	\caption{Two sample images from KITTI and MAPO dataset.}
	
	\label{fig: mapo_img}
\end{figure}

The MAPO dataset is collected along a bike route, making it inherently challenging due to potential occlusion caused by nearby trees. Fig. \ref{fig: mapo_img} shows two images from the MAPO and KITTI datasets, respectively. Table \ref{table: mapo_peformance} shows that MIM-I2P achieves a Top-1 recall rate of 41.8\% using the model trained on sequence 00 of KITTI, validating its strong generalization ability. After the refinement, the recall rate at Top-1 and Top-1\% improves to 57.3\% and 91.4\%, respectively.   

 \begin{table}[t]
	\caption{Performance of MIM-I2P on the MAPO dataset.}
	\vspace{-3mm}
	\label{table: mapo_peformance}
	\begin{center}
		\setlength{\tabcolsep}{1.5mm}{
			\begin{tabular}{c c c c }
				\hline
				&Recall@1 &Recall@1\%  \\
				\hline
				Trained on KITTI & 41.8 & 75.4 \\
				Refined on MAPO & 57.3 & 91.4 \\
				\hline
			\end{tabular}
			\vspace{-5mm}
		}
	\end{center}
\end{table}

\section{Conclusion}
In this work, we propose a cross-modal place recognition method called I2P-Rec that can localize images on large-scale point cloud maps. We leverage the depth estimation algorithms to recover point clouds from images to eliminate the modal difference. We further perform BEV projection for global feature extraction. Although the strategy of our method is simple, it is quite effective and has good generalization ability as the experiments conducted on the KITTI and the MAPO dataset show. In our future work, we will design more effective modality transformation algorithms to boost the performance of our method. We also plan to build a real-time visual localization system working on large-scale point cloud maps.  


\bibliographystyle{IEEEtran}
\bibliography{reference}

\begin{thebibliography}{10}
\providecommand{\url}[1]{#1}
\csname url@rmstyle\endcsname
\providecommand{\newblock}{\relax}
\providecommand{\bibinfo}[2]{#2}
\providecommand\BIBentrySTDinterwordspacing{\spaceskip=0pt\relax}
\providecommand\BIBentryALTinterwordstretchfactor{4}
\providecommand\BIBentryALTinterwordspacing{\spaceskip=\fontdimen2\font plus
\BIBentryALTinterwordstretchfactor\fontdimen3\font minus
  \fontdimen4\font\relax}
\providecommand\BIBforeignlanguage[2]{{%
\expandafter\ifx\csname l@#1\endcsname\relax
\typeout{** WARNING: IEEEtran.bst: No hyphenation pattern has been}%
\typeout{** loaded for the language `#1'. Using the pattern for}%
\typeout{** the default language instead.}%
\else
\language=\csname l@#1\endcsname
\fi
#2}}

\bibitem{cadena2016past}
C.~Cadena, L.~Carlone, H.~Carrillo, Y.~Latif, D.~Scaramuzza, J.~Neira, I.~Reid,
  and J.~J. Leonard, ``Past, present, and future of simultaneous localization
  and mapping: Toward the robust-perception age,'' \emph{IEEE Transactions on
  Robotics}, vol.~32, no.~6, pp. 1309--1332, 2016.

\bibitem{lu2019l3}
W.~Lu, Y.~Zhou, G.~Wan, S.~Hou, and S.~Song, ``{L3-Net}: {Towards} learning
  based {LiDAR} localization for autonomous driving,'' in \emph{IEEE Conference
  on Computer Vision and Pattern Recognition}, 2019, pp. 6389--6398.

\bibitem{feng20192d3d}
M.~Feng, S.~Hu, M.~H. Ang, and G.~H. Lee, ``{2D3D-MatchNet}: Learning to match
  keypoints across {2D} image and {3D} point cloud,'' in \emph{International
  Conference on Robotics and Automation}.\hskip 1em plus 0.5em minus
  0.4em\relax IEEE, 2019, pp. 4790--4796.

\bibitem{chen2022i2d}
K.~Chen, H.~Yu, W.~Yang, L.~Yu, S.~Scherer, and G.-S. Xia, ``{I2D-Loc}: Camera
  localization via image to lidar depth flow,'' \emph{ISPRS Journal of
  Photogrammetry and Remote Sensing}, vol. 194, pp. 209--221, 2022.

\bibitem{angelina2018pointnetvlad}
M.~Angelina~Uy and G.~Hee~Lee, ``{PointNetVLAD}: Deep point cloud based
  retrieval for large-scale place recognition,'' in \emph{IEEE Conference on
  Computer Vision and Pattern Recognition}.\hskip 1em plus 0.5em minus
  0.4em\relax IEEE, 2018, pp. 4470--4479.

\bibitem{liu2019lpd}
Z.~Liu, S.~Zhou, C.~Suo, P.~Yin, W.~Chen, H.~Wang, H.~Li, and Y.-H. Liu,
  ``{LPD-Net}: {3D} point cloud learning for large-scale place recognition and
  environment analysis,'' in \emph{IEEE International Conference on Computer
  Vision}.\hskip 1em plus 0.5em minus 0.4em\relax IEEE, 2019, pp. 2831--2840.

\bibitem{bvmatch2021}
L.~Luo, S.-Y. Cao, B.~Han, H.-L. Shen, and J.~Li, ``{BVMatch}: Lidar-based
  place recognition using bird's-eye view images,'' \emph{IEEE Robotics and
  Automation Letters}, vol.~6, no.~3, pp. 6076--6083, 2021.

\bibitem{du2020dh3d}
J.~Du, R.~Wang, and D.~Cremers, ``{DH3D}: {Deep} hierarchical {3D} descriptors
  for robust large-scale {6-DoF} relocalization,'' in \emph{IEEE Conference on
  Computer Vision and Pattern Recognition}.\hskip 1em plus 0.5em minus
  0.4em\relax Springer, 2020, pp. 744--762.

\bibitem{luo2022lidar}
L.~Luo, S.-Y. Cao, Z.~Sheng, and H.-L. Shen, ``{LiDAR-based} global
  localization using histogram of orientations of principal normals,''
  \emph{IEEE Transactions on Intelligent Vehicles}, vol.~7, no.~3, pp.
  771--782, 2022.

\bibitem{2012Bags}
D.~Galvez-Lpez and J.~D. Tardos, ``Bags of binary words for fast place
  recognition in image sequences,'' \emph{IEEE Transactions on Robotics},
  vol.~28, no.~5, pp. 1188--1197, 2012.

\bibitem{arandjelovic2016netvlad}
R.~Arandjelovic, P.~Gronat, A.~Torii, T.~Pajdla, and J.~Sivic, ``{NetVLAD}:
  {CNN} architecture for weakly supervised place recognition,'' in \emph{IEEE
  Conference on Computer Vision and Pattern Recognition}, 2016, pp. 5297--5307.

\bibitem{cattaneo2020global}
D.~Cattaneo, M.~Vaghi, S.~Fontana, A.~L. Ballardini, and D.~G. Sorrenti,
  ``Global visual localization in lidar-maps through shared {2D-3D} embedding
  space,'' in \emph{IEEE International Conference on Robotics and
  Automation}.\hskip 1em plus 0.5em minus 0.4em\relax IEEE, 2020, pp.
  4365--4371.

\bibitem{godard2019digging}
C.~Godard, O.~Mac~Aodha, M.~Firman, and G.~J. Brostow, ``Digging into
  self-supervised monocular depth estimation,'' in \emph{IEEE Conference on
  Computer Vision and Pattern Recognition}, 2019, pp. 3828--3838.

\bibitem{chang2018pyramid}
J.-R. Chang and Y.-S. Chen, ``Pyramid stereo matching network,'' in \emph{IEEE
  Conference on Computer Vision and Pattern Recognition}, 2018, pp. 5410--5418.

\bibitem{xie2023darkmim}
X.~Zhenda, G.~Zigang, H.~Jingcheng, Z.~Zheng, H.~Han, and C.~Yue, ``Revealing
  the dark secrets of masked image modeling,'' \emph{arXiv preprint
  arXiv:2205.13543}, 2022.

\bibitem{luo2023}
L.~Lun, Z.~Shuhang, L.~Yixuan, F.~Yongzhi, Y.~Beinan, S.~Cao, and H.-L. Shen,
  ``{BEVPlace}: {Learning LiDAR-based} place recognition using bird's eye view
  images,'' \emph{arXiv preprint arXiv:2302.14325}, 2023.

\bibitem{kitti}
A.~Geiger, P.~Lenz, and R.~Urtasun, ``Are we ready for autonomous driving? the
  {KITTI} vision benchmark suite,'' in \emph{IEEE Conference on Computer Vision
  and Pattern Recognition}, 2012, pp. 3354--3361.

\bibitem{2004Distinctive}
D.~Lowe, ``Distinctive image features from scale-invariant keypoints,''
  \emph{International Journal of Computer Vision}, vol.~20, pp. 91--110, 2004.

\bibitem{rublee2011orb}
E.~Rublee, V.~Rabaud, K.~Konolige, and G.~Bradski, ``{ORB}: An efficient
  alternative to {SIFT} or {SURF},'' in \emph{IEEE Conference on Computer
  Vision and Pattern Recognition}.\hskip 1em plus 0.5em minus 0.4em\relax IEEE,
  2011, pp. 2564--2571.

\bibitem{lazebnik2006beyond}
S.~Lazebnik, C.~Schmid, and J.~Ponce, ``Beyond bags of features: Spatial
  pyramid matching for recognizing natural scene categories,'' in \emph{IEEE
  Conference on Computer Vision and Pattern Recognition}, vol.~2.\hskip 1em
  plus 0.5em minus 0.4em\relax IEEE, 2006, pp. 2169--2178.

\bibitem{2017ORB}
R.~Mur-Artal, J.~M.~M. Montiel, and J.~D. Tardos, ``{ORB-SLAM}: A versatile and
  accurate monocular {SLAM} system,'' \emph{IEEE Transactions on Robotics},
  vol.~31, no.~5, pp. 1147--1163, 2017.

\bibitem{pronobis2006discriminative}
A.~Pronobis, B.~Caputo, P.~Jensfelt, and H.~I. Christensen, ``A discriminative
  approach to robust visual place recognition,'' in \emph{IEEE/RSJ
  International Conference on Intelligent Robots and Systems}.\hskip 1em plus
  0.5em minus 0.4em\relax IEEE, 2006, pp. 3829--3836.

\bibitem{lowry2015visual}
S.~Lowry, N.~S{\"u}nderhauf, P.~Newman, J.~J. Leonard, D.~Cox, P.~Corke, and
  M.~J. Milford, ``Visual place recognition: {A} survey,'' \emph{IEEE
  Transactions on Robotics}, vol.~32, no.~1, pp. 1--19, 2015.

\bibitem{angeli2008fast}
A.~Angeli, D.~Filliat, S.~Doncieux, and J.-A. Meyer, ``Fast and incremental
  method for loop-closure detection using bags of visual words,'' \emph{IEEE
  Transactions on Robotics}, vol.~24, no.~5, pp. 1027--1037, 2008.

\bibitem{hausler2021patch}
S.~Hausler, S.~Garg, M.~Xu, M.~Milford, and T.~Fischer, ``{Patch-NetVLAD:
  Multi-scale} fusion of locally-global descriptors for place recognition,'' in
  \emph{IEEE Conference on Computer Vision and Pattern Recognition}, 2021, pp.
  14\,141--14\,152.

\bibitem{qi2017pointnet}
C.~R. Qi, H.~Su, K.~Mo, and L.~J. Guibas, ``{PointNet}: Deep learning on point
  sets for {3D} classification and segmentation,'' in \emph{IEEE Conference on
  Computer Vision and Pattern Recognition}, 2017, pp. 652--660.

\bibitem{zhang2019pcan}
W.~{Zhang} and C.~{Xiao}, ``{PCAN}: {3D} attention map learning using
  contextual information for point cloud based retrieval,'' in \emph{IEEE
  Conference on Computer Vision and Pattern Recognition}, 2019, pp.
  12\,428--12\,437.

\bibitem{soe}
Y.~Xia, Y.~Xu, S.~Li, R.~Wang, J.~Du, D.~Cremers, and U.~Stilla, ``{SOE-Net}: A
  self-attention and orientation encoding network for point cloud based place
  recognition,'' in \emph{IEEE Conference on Computer Vision and Pattern
  Recognition}, 2021, pp. 11\,343--11\,352.

\bibitem{mickloc3dv2}
{J. Komorowski}, ``Improving point cloud based place recognition with
  ranking-based loss and large batch training,'' in \emph{International
  Conference on Pattern Recognition}.\hskip 1em plus 0.5em minus 0.4em\relax
  IEEE, 2022, pp. 3699--3705.

\bibitem{kim2018scan}
G.~Kim and A.~Kim, ``{Scan Context}: Egocentric spatial descriptor for place
  recognition within {3D} point cloud map,'' in \emph{IEEE International
  Conference on Intelligent Robots and Systems}.\hskip 1em plus 0.5em minus
  0.4em\relax IEEE, 2018, pp. 4802--4809.

\bibitem{chen2020overlapnet}
X.~Chen, T.~L{\"a}be, A.~Milioto, T.~R{\"o}hling, O.~Vysotska, A.~Haag,
  J.~Behley, C.~Stachniss, and F.~Fraunhofer, ``{OverlapNet}: Loop closing for
  {LiDAR-based SLAM},'' in \emph{Proc. of Robotics: Science and Systems}, 2020.

\bibitem{OverlapTransformer}
J.~Ma, J.~Zhang, J.~Xu, R.~Ai, W.~Gu, and X.~Chen, ``{OverlapTransformer}: An
  efficient and yaw-angle-invariant transformer network for {LiDAR}-based place
  recognition,'' \emph{IEEE Robotics and Automation Letters}, vol.~7, no.~3,
  pp. 6958--6965, 2022.

\bibitem{bernreiter2021spherical}
L.~Bernreiter, L.~Ott, J.~Nieto, R.~Siegwart, and C.~Cadena, ``Spherical
  multi-modal place recognition for heterogeneous sensor systems,'' in
  \emph{IEEE International Conference on Robotics and Automation}.\hskip 1em
  plus 0.5em minus 0.4em\relax IEEE, 2021, pp. 1743--1750.

\bibitem{esteves2018learning}
C.~Esteves, C.~Allen-Blanchette, A.~Makadia, and K.~Daniilidis, ``Learning
  {SO}(3) equivariant representations with spherical {CNNs},'' in
  \emph{European Conference on Computer Vision}, 2018, pp. 52--68.

\bibitem{cohen2018spherical}
T.~S. Cohen, M.~Geiger, J.~K{\"o}hler, and M.~Welling, ``Spherical {CNN},''
  \emph{arXiv preprint arXiv:1801.10130}, 2018.

\bibitem{monosurvey}
X.~Dong, M.~A. Garratt, S.~G. Anavatti, and A.~H. A., ``Towards real-time
  monocular depth estimation for robotics: A survey,'' \emph{arXiv preprint
  arXiv:2111.08600}, 2021.

\bibitem{he2016deep}
K.~He, X.~Zhang, S.~Ren, and J.~Sun, ``Deep residual learning for image
  recognition,'' in \emph{IEEE Conference on Computer Vision and Pattern
  Recognition}, 2016, pp. 770--778.

\bibitem{shrivastava2016training}
A.~Shrivastava, A.~Gupta, and R.~Girshick, ``Training region-based object
  detectors with online hard example mining,'' in \emph{IEEE Conference on
  Computer Vision and Pattern Recognition}, 2016, pp. 761--769.

\bibitem{cheng2020stereo}
X.~Cheng, Y.~Zhong, M.~Harandi, Y.~Dai, X.~Chang, H.~Li, T.~Drummond, and
  Z.~Ge, ``Hierarchical neural architecture search for deep stereo matching,''
  in \emph{Advances in Neural Information Processing Systems}, vol.~33, 2020,
  pp. 22\,158--22\,169.

\end{thebibliography}

\end{document}